\documentclass[utf8]{FrontiersinHarvard} 

\usepackage{url,hyperref,lineno,microtype,subcaption}
\usepackage[onehalfspacing]{setspace}
\usepackage{svg}
\usepackage{float}
\usepackage{tabularx}
\usepackage[margin=1in]{geometry}
\usepackage{array}
\usepackage{microtype}
\usepackage{cleveref}
\usepackage{subfiles}
\usepackage{booktabs}

\def\keyFont{\fontsize{8}{11}\helveticabold }
\def\firstAuthorLast{Dyck {et~al.}} 
\def\Authors{Leonie Dyck\,$^{1}$, Aiko Galetzka\,$^{1}$, Maximilian Noller\,$^{1}$, Anna-Lena Rinke\,$^{1}$, Jutta Bormann\,$^{2}$, Jekaterina Miller\,$^{2}$, Michelle Hochbaum\,$^{2}$, Julia Siemann\,$^{2}$, J\"{o}rdis Alboth\,$^{2}$, Andre Berwinkel\,$^{2}$, Johanna Luz\,$^{2}$, Britta Kley-Zobel\,$^{3}$, Marcine Cyrys\,$^{3}$, Nora Fl\"{o}ttmann\,$^{2}$,  Ariane Vogeler\,$^{2}$, Mariia Melnikova\,$^{2}$, Ira-Katharina Petras\,$^{2\footnote{now Faculty of Medicine, RWTH Aachen University}}$, Michael Siniatchkin\,$^{2\footnotemark[2]}$, Winfried Barthlen\,$^{2}$ and Anna-Lisa Vollmer\,$^{1,*}$}
\def\Address{$^{1}$Medical School OWL, Bielefeld University, 33615 Bielefeld,
Germany \\
$^{2}$Evangelisches Klinikum Bethel gGmbH, 33617 Bielefeld, Germany \\
$^{3}$Hand \& Fuß Physiotherapie, 33719 Bielefeld, Germany}
\def\corrAuthor{Anna-Lisa Vollmer}

\def\corrEmail{anna-lisa.vollmer@uni-bielefeld.de}

\begin{document}
\onecolumn
\firstpage{1}

\title {Interprofessional and Agile Development of \emph{Mobirobot}: A Socially Assistive Robot for Pediatric Therapy Across Clinical and Therapeutic Settings} 

\author[\firstAuthorLast ]{\Authors} 
\address{} 
\correspondance{} 

\extraAuth{}

\maketitle

\begin{abstract}
\textbf{Introduction:} Socially assistive robots hold promise for enhancing therapeutic engagement in paediatric clinical settings. However, their successful implementation requires not only technical robustness but also context-sensitive, co-designed solutions. This paper presents \textit{Mobirobot}, a socially assistive robot developed to support mobilisation in children recovering from trauma, fractures, or depressive disorders through personalised exercise programmes.

\textbf{Methods:} An agile, human-centred development approach guided the iterative design of \textit{Mobirobot}. Multidisciplinary clinical teams and end users were involved throughout the co-development process, which focused on early integration into real-world paediatric surgical and psychiatric settings. The robot, based on the NAO platform, features a simple setup, adaptable exercise routines with interactive guidance, motivational dialogue, and a graphical user interface (GUI) for monitoring and no-code system feedback.

\textbf{Results:} Deployment in hospital environments enabled the identification of key design requirements and usability constraints. Stakeholder feedback led to refinements in interaction design, movement capabilities, and technical configuration. A feasibility study is currently underway to assess acceptance, usability, and perceived therapeutic benefit, with data collection including questionnaires, behavioural observations, and staff--patient interviews.

\textbf{Discussion:} \textit{Mobirobot} demonstrates how multiprofessional, stakeholder-led development can yield a socially assistive system suited for dynamic inpatient settings. Early-stage findings underscore the importance of contextual integration, robustness, and minimal-intrusion design. While challenges such as sensor limitations and patient recruitment remain, the platform offers a promising foundation for further research and clinical application.

\textbf{Conclusion:} By aligning participatory development with therapeutic objectives, \textit{Mobirobot} advances the field of paediatric socially assistive robotics. Its design principles and identified feedback themes may serve as valuable guidelines for future deployments in digital health.

\tiny
 \keyFont{ \section{Keywords:} participation, interprofessional participatory design, co-development, agile development, sports therapy, assistive robot, assistance system, human-robot interaction} 
\end{abstract}

\section{Introduction}
Our work focuses the employment of a socially assistive robot (SAR) for mobilization of children and adolescents in two main use cases: supporting sports therapy for psychiatric patients with depression or anxiety disorders (age 8 to 18), and supporting somatic physiotherapy for patients after abdominal trauma or fractures of the lower limbs (age 4 to 16).

Psychiatric disorders among adolescents and children are a growing concern for public healthcare systems worldwide \citep{bitsko_epidemiology_2018,keyes_recent_2019,zablotsky_prevalence_2019,xu_twenty-year_2018} with a lifetime prevalence rate of $16,9\%$ \citep{klipker_psychische_2018}. 
For this reason, going beyond established first-line psychotherapy and pharmacotherapy, sports and exercise therapy plays an important role in the treatment of mental health disorders. Its effectiveness has been confirmed in several meta-analyses and systematic reviews for adults \citep{cooney_exercise_2013,krogh_effect_2011,schuch_exercise_2016,strohle_sports_2019} as well as children and adolescents with psychiatric disorders \citep{ash_physical_2017}.
In addition, studies also suggest that the full potential of sports and exercise therapy is not yet being exploited in many clinics \citep{brehm2019angebot} and that the recommendations of the WHO of 60 minutes moderate or intense physical activity per day can only be achieved by $26\%$ of adolescents \citep{robert_koch-institut_korperliche_2018}. For this reason, SARs have the potential to support sports and exercise therapy with additional offers, motivate and increase the participation in sports and exercise interventions, to help to build up a physically active lifestyle and even to relieve the hospital staff.

In their work ''Physiotherapy in intensive care medicine", G{\"a}rtner and Roth emphasize that physiotherapy after surgery plays a key role in preventing complications such as respiratory infections, circulatory problems and movement disorders \citep{Gartner_Roth_2000}. This is particularly true for children, who often have reduced physical activity after surgery, which increases the risk of complications. Early and regular physiotherapy mobilizations can help to minimize the risk of such complications and shorten the overall recovery time \citep{pulido2017evaluating,mataric2007socially}. However, the availability of experienced pediatric physical and sports therapists is limited, and their services are often costly.

The integration of social assistive robotics has been demonstrated to enhance social interaction and positive affect in older adults, reduce anxiety and worry in children, improve mood and stress, and reduce pain across the lifespan \citep{NICHOL2024104730}.

To address these use cases, we propose the integration of humanoid robots to support therapists in mobilizing children in both, pediatric wards and outpatient settings. These robots can take on tasks related to movement demonstration and motivation, leveraging children’s generally positive attitudes toward robots \citep{dawe2019can}. The embodiment of robots proves advantageous in demonstrating exercises, fostering motivation, and enhancing compliance \citep{pulido2017evaluating}.
Nevertheless, the effectiveness of SAR in these roles hinges on their successful integration into daily clinical and outpatient routines. Their acceptance, compliance, and engagement by children, parents, and medical caregivers are critical factors for success \citep{aguiar2021barriers,klemme2021multi}. By combining the therapeutic benefits of exercise with the motivational advantages of SARs, there is a promising opportunity to improve both mental health outcomes and physical rehabilitation in pediatric populations \citep{dawe2019can}.

\begin{figure}
    \centering
    \includegraphics[width=\columnwidth]{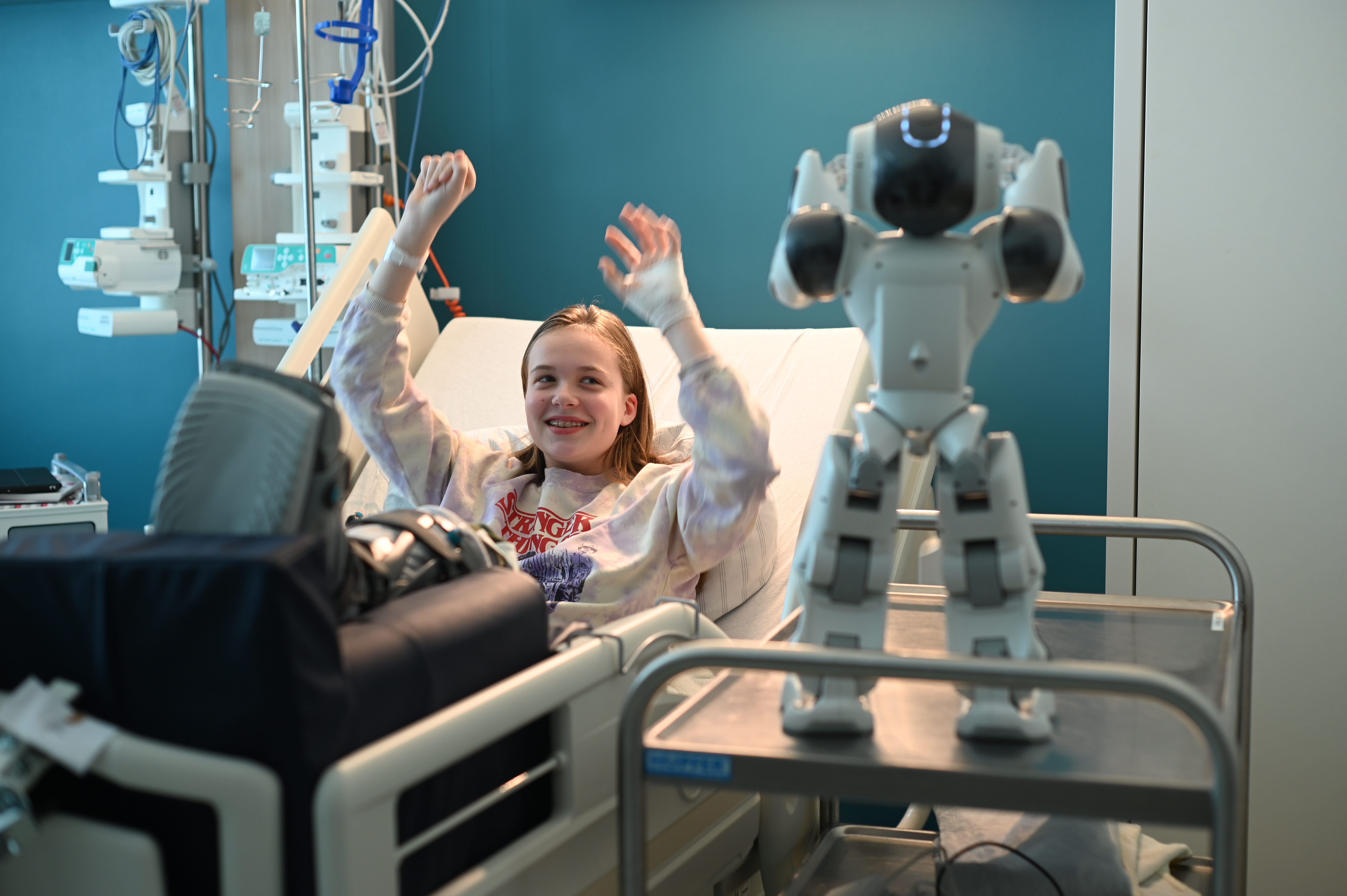}
    \caption{Mobirobot during physiotherapy with a patient in the InPT Setting.}
    \label{fig:kc}
\end{figure}

\section{Related Work}

\subsection{Socially Assistive Robots for Supporting Pediatric Therapy}
New assistive technologies, such as SARs, have the potential to assist professionals in treating patients and to expand the range of treatments available. 

SARs are robots that are specifically designed to interact with humans by adapting to human behaviors and rules \citep{huber_designing_2014}. Individual studies report robotic use to assist sports and physical therapists, especially with children and adolescents with infantile cerebral palsy \citep{abdul_rahman_use_2015,abdul_malik_potential_2016,kozyavkin_humanoid_2014}. Other uses of SARs include for example sports therapy for adolescents with obesity \citep{swift-spong_comparing_2016}, exercise therapy for elderly patients with dementia \citep{schrum_humanoid_2019,cruz-sandoval_towards_2018}, or motor and gait training for patients in rehabilitation after suffering a stroke \citep{lee_towards_2020,kim_robot-assisted_2019}.

SARs are also used in child and adolescent psychiatry for therapeutic purposes, especially in the treatment of autistic disorders \citep{doi:10.1126/scirobotics.aat7544}
SARs help children with social communication impairment by serving as play partners transitioning to human contact. Children with autism sprectrum disorder (ASD) are often overwhelmed with the complexity of human facial expressions and gestures, making human contact unbearable for them \citep{robinson_psychosocial_2019}. Although humanoid robots resemble humans, their simplified appearance and predictable behavior make them more approachable for autistic children. SARs offer a gentle introduction to human-like interaction, allowing children to explore social cues at their own pace, experiment with different behavioral responses, and gradually build confidence in engaging with real people \citep{cabibihan_why_2013}. 

The most commonly used humanoid robots for such socially assistive tasks are the NAO and Pepper robots. The NAO robot is an autonomous humanoid robot that can walk, talk, and recognize sounds. It is especially suited for physical movement generation because of its robustness and motion capabilities. Pepper is also an autonomous humanoid robot that can interact with humans both verbally and physically. This is done, for example, by changing the color of its eyes and the tone of its voice \citep{littler_reducing_2021}. As mentioned above, there are many applications for SARs in healthcare. For example, first attempts are being made to use humanoid robots in various care settings for children undergoing oncological treatment \citep{alemi2016clinical}. In this case, the goal of SARs is to provide support in coping with the current situation and to actively counteract the children's feelings of fear, anger and depression \citep{beyer2020effects}. 

But SAR use in child and adolescent psychiatric patients with other medical conditions has not been adequately explored, despite promising results indicating that SARs have a positive impact on psychological well-being in children and adolescents \citep{liverpool_engaging_2020,kazdin_annual_2019,robinson_psychosocial_2019,littler_reducing_2021,moerman_social_2019}.

There is also limited research on the use of SARs in supporting surgical patients.
A recent study indicates that interactive robots can positively influence postoperative mobilization in children undergoing day surgery \citep{topccu2023effect}. 
However, current research has not yet explored the benefits of SARs for promoting mobilization and recovery in children and adolescents following abdominal surgeries or fractures of the lower extremities.

\subsection{Participatory Development in Healthcare Application}
In the field of health sciences, it is increasingly recognized that a key to enhancing the effectiveness and sustainability of interventions is the comprehensive inclusion of affected individuals along with their needs \citep{larsson2018children}. Demonstrated advantages of enhanced user participation in design include: (a) enhanced quality of the system due to more precise and effective collection of user requirements, (b) an increased probability of integrating desired features while excluding costly, unwanted features during the design phase, (c) elevated user acceptance of the system, which was developed with substantial user input, (d) better comprehension of the system by end users, resulting in reduced training requirements and fewer operational problems, and (e) an increased role for users in the decision-making processes within their organizations \citep{kujala2003user}.
The scientific approach of participatory research thus pursues the objective to gain new insights by intertwining experiential knowledge (e.g., from daily life) with scientific knowledge in a collaborative learning process \citep{percy2014reclaiming}, to collectively generate new knowledge by continuing the process in a spiral until all participants feel benefit \citep{weiss2022robots,lee2017steps}.
In the field of human-robot interaction (HRI), participatory design (PD) is gaining popularity and is recognized for its capacity to collaboratively create impactful technologies \citep{harrington2019deconstructing}. Central to PD is the principle of mutual learning between researchers (designers) and non-professionals (experts in their own experiences), aiming to diminish social hierarchies by recognizing each party's agency in decision-making \citep{bannon2012design,lee2017steps}. PD is particularly effective for addressing real-world challenges that require the joint efforts of designers, developers, and subject-matter experts to devise solutions that are both relevant and meaningful \citep{von2009democratizing}.
\citet{merkel2019participatory} reviewed participatory development approaches in gereontechnology. Referring to \citet{panek2017levels} and utilizing the ladder of participation model \citep{arnstein1969ladder}, they categorized four degrees of involvement in participatory design. These are classified as

\textbf{No Involvement}: Anticipation of users' needs and preferences is based on assumptions or existing literature, with the use of personas.

\textbf{Low Involvement}: Involvement occurs through surveys or observational studies to gather preferences from users, often facilitated by an institution like a user's organization. (Ethnographic/``In-the-Wild'' Studies and User-Centred Design, cf. \citep{lee2017steps,winkle2021leador})

\textbf{Medium Involvement}: Involvement includes participation in specific stages of the design process, such as the evaluation of a prototype, where users can directly and actively influence a critical phase of the design.

\textbf{Full Involvement}: Users are involved as equal partners throughout all stages of the design process, with the ability to actively influence every aspect, including the decision to conclude the process. (Participatory Design, cf. \citep{lee2017steps,winkle2021leador})

Our co-development process has set out with a mutual need in employment settings and features the full involvement of the multiprofessional stakeholder group of different settings, continuous integration of requirements and the medium involvement of patients in the in-situ testing of prototypes. We combined participatory development with agile software development methods informed by Feature Driven Development \citep{abrahamsson2017agile}.

\section{Methods}
\subsection{Deployment Settings}
\begin{figure}
    \centering
    \includegraphics[width=\columnwidth]{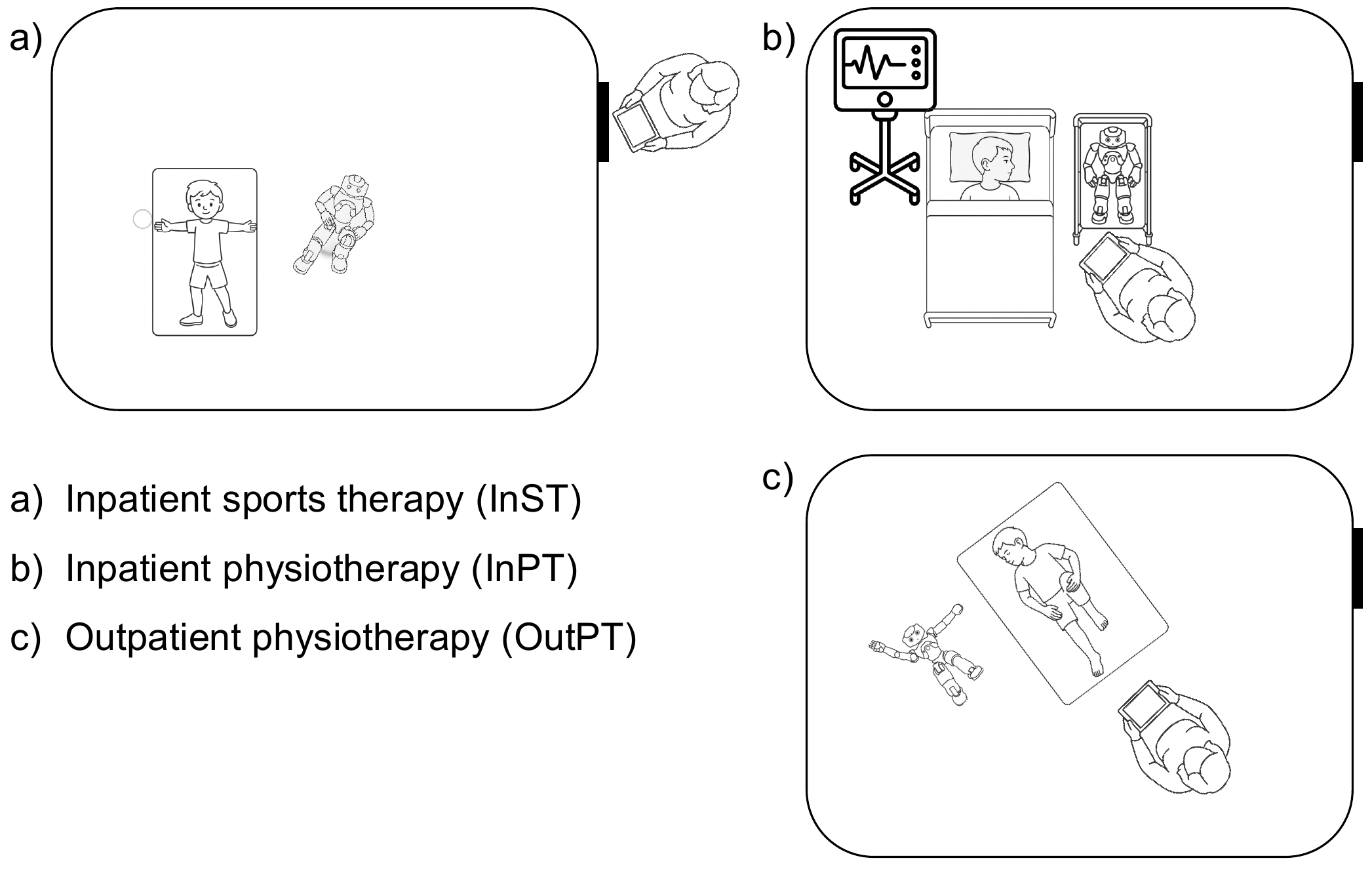}
    \caption{Schematic depiction of the setups in the three settings a)-c).}
    \label{fig:setups}
\end{figure}

The pediatric deployment settings of Mobirobot can be classified along two key dimensions: clinical context, distinguishing between inpatient and outpatient care, and therapeutic focus, differentiating between physiotherapy and sports-based therapy. The setup differed according to the setting as described below and depicted in Fig. \ref{fig:setups}.

\subsubsection{Paediatric inpatient sports therapy for psychiatric patients (InST)}
The clinical inpatient service context for sports therapy is represented by the University Clinic for Child and Adolescent Psychiatry at the Bethel Protestant Hospital (EvKB). 
The regular voluntary sports program for children offers a group exercise session every Monday. In addition to these weekly group sessions, an innovative option is envisioned: children can sign up to borrow an exercise robot. They can then take it to the dedicated gym and participate in individual training sessions led by the robot.
\subsubsection{Paediatric inpatient physiotherapy for abdominal surgical patients (InPT)}
The other clinical inpatient setting is the Clinic for Pediatric and Adolescent Surgery and Pediatric Urology at EvKB 
with a focus on physiotherapy. The clinic is located in the new Bethel Children's Center.
Patients undergoing abdominal surgery receive daily physiotherapy sessions from the first to the third postoperative day, each lasting about 30 minutes. The actual duration of each session is tailored to the individual's condition and well-being. Patients at EvKB with fractures of the lower extremities also receive physiotherapy twice weekly for approximately 30 minutes in their respective rooms.
The employment of the robot in the physiotherapy program serves two main purposes: it is used after abdominal surgeries or trauma and following fractures of the lower extremities. The robot should assist in achieving rapid mobilization, which is crucial for fast recovery, by enhancing motivation and demonstrating exercises as a peer.
Each physiotherapy session is conducted under the supervision of a qualified physiotherapist, ensuring that children are never left alone with the robot. This supervision guarantees that the exercises are performed correctly, preventing any potential incorrect execution of the exercises.
\subsubsection{Paediatric outpatient physiotherapy for patients with lower leg fracture (OutPT)}
After discharge, all patients with fractures of the lower extremities are referred to the outpatient physiotherapy practice ``Hand und Fuß,'' (Hand and Foot), where their physiotherapy continues in the treatment room. This ensures a seamless transition from hospital-based care to outpatient therapy, maintaining consistency in treatment and aiding in the patients' recovery.
The practice treats infants, children, and adolescents with a variety of illnesses using common child therapy concepts with a total of ten staff members. In this setting also each physiotherapy session with the robot is conducted under the supervision of a qualified physiotherapist.

\subsection{Equipment}

\subsubsection{Robot Platform}

The socially assistive robot used in this study is the humanoid robot NAO6 developed by Aldebaran Robotics. We employ NAO because of its embodiment, especially in terms of its friendly and approachable appearance and its ability to perform a variety of exercises (\cite{naosurvey}).

Due to its compact size of $58cm$ ($22.8in$), NAO is particularly suitable for children. Additionally, it can be placed on an elevated platform, allowing children with mobility impairments who are confined to sitting or lying down to participate. Additionally, previous research has demonstrated that NAO is well-received by children and can enhance the speed of learning and comprehension. \cite{yousif2018questionnaire}

\subsubsection{Software Prototype}
\label{sec:prototype}

\begin{figure}
    \centering
    \includegraphics[width=.5\columnwidth]{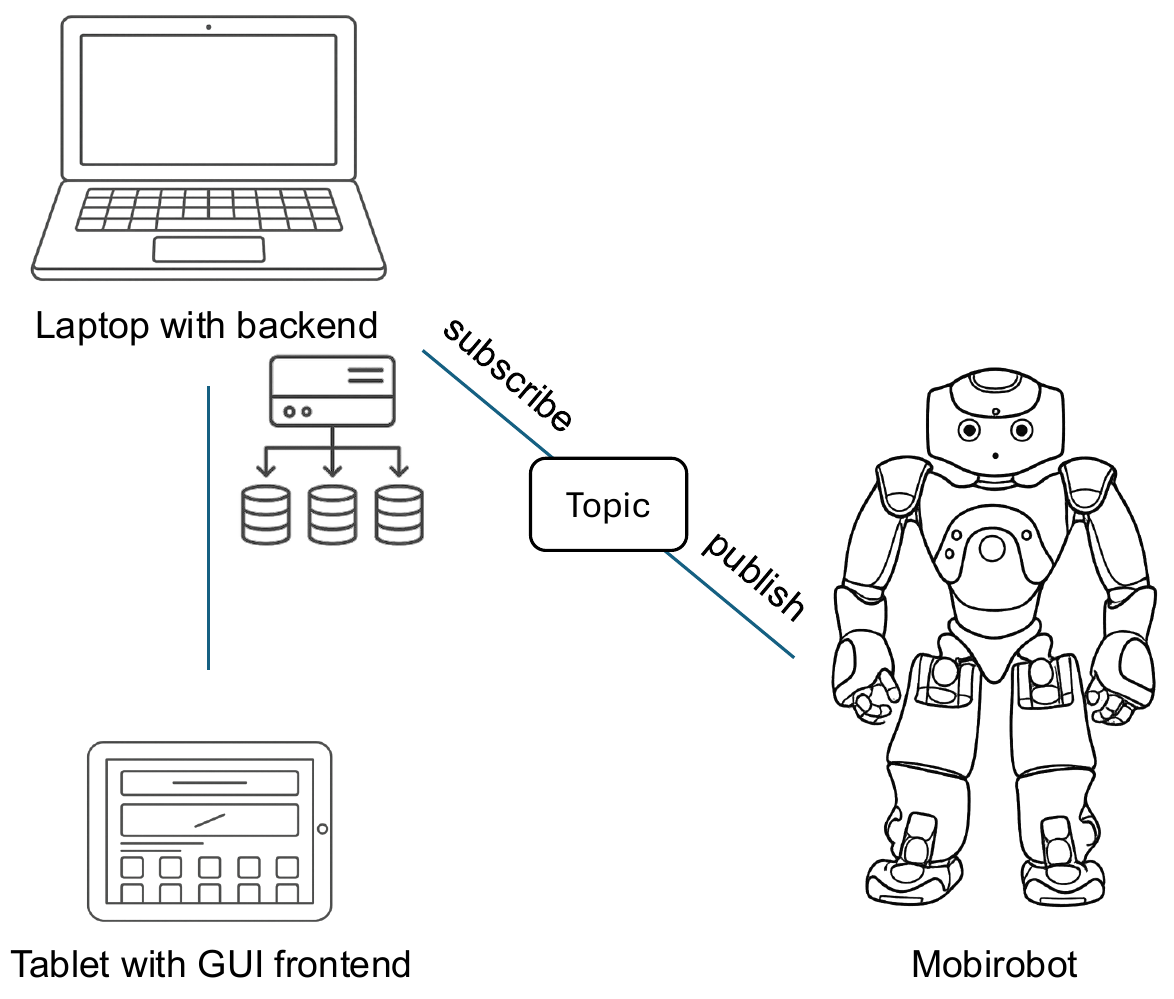}
    \caption{Overview of communication between the different system components. The connections represent local Wi-Fi links initiated by the laptop.}
    \label{fig:system}
\end{figure}

The initial software consisted of a main program running on a laptop and a GUI which can be accessed on a tablet. The prototype was employed in a pilot study run at the InST setting.
The GUI communicates with the backend via a REST-API, as well as a WebSocket-API.
The WebSocket-API is used for displaying information on the real-time state of the robot, e.g. the battery charge or the joints' temperatures. The displayed information are a result of the communication between NAO and the backend via a publisher-subscriber-protocol. NAO publishes certain information on defined topics, e.g. ``HotDeviceDetected'', that the backend can subscribe to. Once the backend receives the information from NAO, it can forward them to the frontend via the WebSocket-API. 
The REST-API is used to send information to the backend that do not lead to a real-time change in the GUI. An overview of this communication between the different components can be found in Fig. \ref{fig:system}.
For the prototype, the GUI was composed of different parts. The first part allowed the therapist to start, stop or pause the program. The program is a sequence of physical exercises that Mobirobot would present one after the other to the participant. This program with the included exercises was designed in collaboration with the sports therapist and fixed. 

During the training with Mobirobot, the participants wore smart watches that recorded their heart rate. The recorded heart rate was displayed in part two of the GUI. If the heart rate got too high or too low (crossing respective thresholds), Mobirobot would respond by telling the participant not to train too hard or to pick up the pace a little bit, respectively.

Part three of the GUI provided an overview of different settings of the robot. The battery status of the robot can be seen at the top of the section, while the volume could be changed at the bottom of the section. The space between these two settings belonged to the speech recognition that was used at the time. After explaining an exercise, Mobirobot would ask the participant whether they had understood the explanation. The participants could answer with ``yes'' or ``no''. The recognized responses from participants were intended to be displayed on the right side of the screen. However, this setup proved ineffective due to the insufficient accuracy of NAO's built-in speech recognition. As a result, the feature was not actively used and answers were manually chosen by the therapist instead.

 

\subsubsection{Exercise program prototype}
\label{sec:exptototype}

The prototype program for robot-assisted sports therapy was developed by an interprofessional team from the Bethel Protestant Hospital (EvKB) and Bielefeld University. According to the literature, especially moderate to vigorous exercise training appears to improve anxiety, depression and impulsivity \citep{sampasa-kanyinga_combinations_2020,kamijo_effects_2019}. Current studies suggest, that especially high-intensity-interval training (HIIT) is appropriate for people with severe mental illness, improve cardio-respiratory fitness and depression and is as feasible as moderate intensity training \citep{korman_high_2020,costigan_high-intensity_2015}.  
Thus, we compiled a list of exercises for HIIT, which was adapted to the capacities of the robot after a thorough review. The review process was based on the parameters of practicability (including understandability by the target group), feasibility, and replicability. The process is shown in the Supplementary Materials File 1. First, exercises were selected, classified and programmed for Mobirobot (step 1).
This step was followed by a first proof of feasibility (internal test run; step 2), after which 18 of 26 exercises remained for employment in the InST setting. The other exercises did not meet the criteria of practicability, feasibility, and replicability. After the pilot study in the Psychiatry, 18 exercises (14 strength exercises and 4 stretching exercises) were subjected to a second proof of feasibility (internal test run after the pilot study (step 3) and were deemed practical, feasible and replicable. These exercises will be used for the main study.

A list of motivational phrases was compiled for Mobirobot, which it randomly picked and uttered during exercises. 
For each exercise, an explanation together with a demonstration was added. Whenever Mobirobot starts a new exercise, it first demonstrates what this next exercise will look like and how it will be done, while also explaining what it is doing. After the demonstration and explanation, the robot performs the exercise together with the patient, which has been found to increase motivation \citep{schneider2016exercising}.

The developed final robot-assisted HIIT training program consists of a structured regime divided into three phases: warm-up, circuit training, and stretching exercises. The warm-up phase, lasting five minutes, was initially conducted using an ergometer. However, the pilot study revealed that some children were too small to use the ergometer effectively, prompting the development of alternatives (as the buttons described in Section~\ref{sec:motivation}). The core component of the training is a 20-minute circuit training session, consisting of four stations. Each station includes four strength exercises, with each exercise being performed for one minute, followed by a 30-second rest period. The exercises include boxing, squads, lunges, Russian twists and sit-ups along with other abdominal exercises (such as dead bug, jackknife, and leg raises), as well as the quadruped position with alternating arm and leg lifts, Superman and plank. 
The training program ends with a five-minute cool-down, during which standardized stretching exercises are performed. Each stretch lasts one minute and includes exercises to stretch the neck, torso, calves, sides, and thighs.

\subsection{Co-development Process}

\begin{figure}[htbp]
  \centering
    \includegraphics[width=\textwidth]{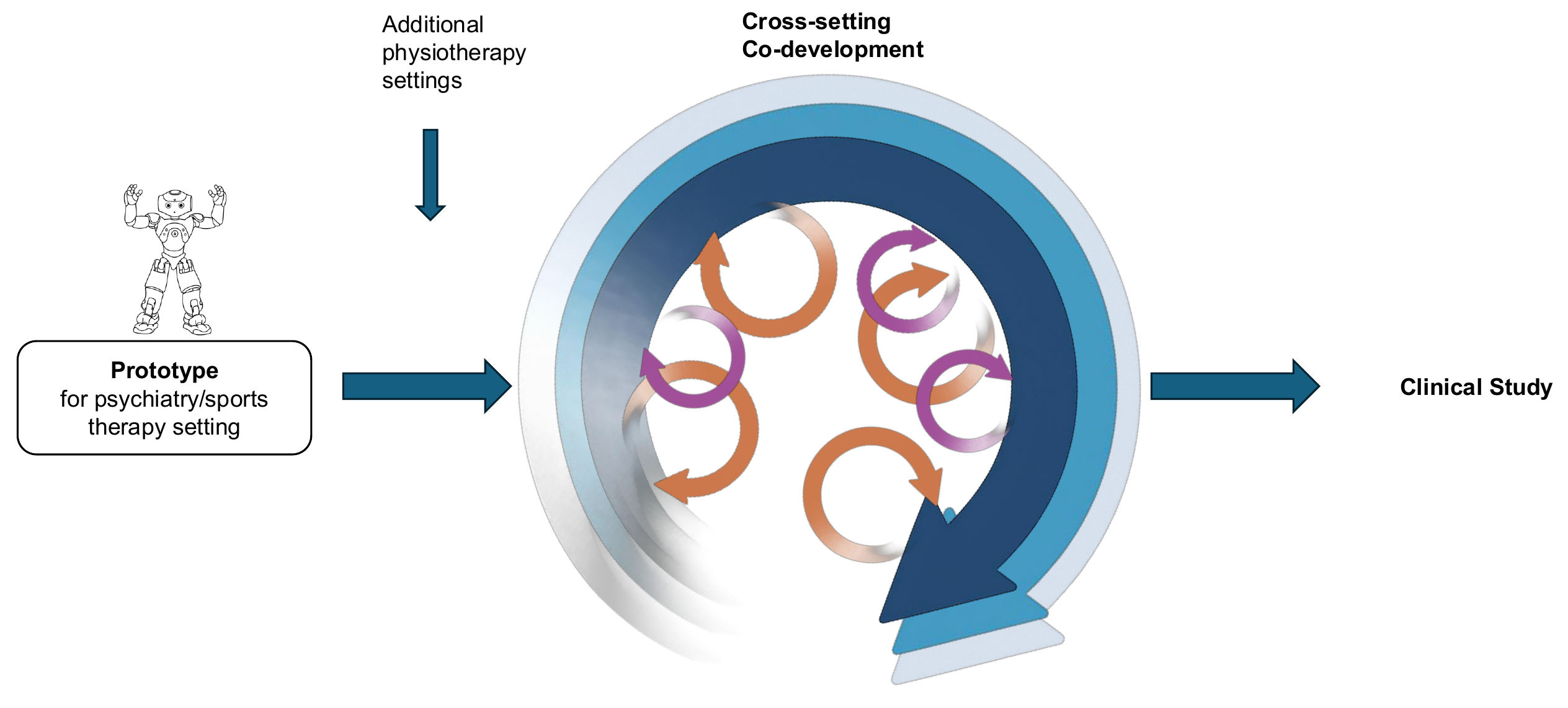}
  \caption{Schematic representation of the iterative, co-development process, illustrating multilevel feedback loops at different scales. Small purple loops depict rapid exchanges on the exercise program via online meetings, email, and file sharing. Medium orange loops represent sprint meetings within the development team, used to prioritize and implement changes. Large loops correspond to full-team meetings featuring system demonstrations and structured feedback sessions as well as engagement activities with patients and clinical staff, including introductory workshops and deployment site visits.}
  \label{fig:cycle}
\end{figure}

The system was developed through an interdisciplinary and interprofessional co-development process over the course of three years across the three deployment settings, involving expertise from computer science, medicine, psychotherapy, psychology, physiotherapy, sports science, and nursing. Specifically, the team included four computer scientists, two medical doctors, one psychotherapist, one psychologist, four physiotherapists, one sports therapist, and one study nurse. The initial prototype was created in response to a clearly articulated clinical need: to increase opportunities for physical activity among inpatients in a child and adolescent psychiatric clinic (cf. left loop in Fig.~\ref{fig:cycle}).

Rather than following a rigid design protocol, the co-development process was contextualized and tailored to the specific requirements of the deployment environments and target users. The development trajectory was shaped by continuous iteration, stakeholder feedback, and real-world testing, in line with principles of human-centered design and agile software development.

The overall co-development encompassed three primary components: the core software of the assistive robot including a graphical user interface (Sections~\ref{sec:robot_software} and \ref{sec:prototype}), the robot-supported exercise program (Section~\ref{sec:exprog}). The evolving co-development process is summarized in Fig.~\ref{fig:cycle}.

\subsubsection{Collaborative Structures and Iterative Workflows}

Monthly full-team meetings served as central touchpoints, combining organizational updates with hands-on workshops. During these sessions, the current system state was demonstrated, and structured feedback was collected from all professional groups. This feedback was prioritized and integrated into the next development cycle.

In parallel, the core developer team (computer scientists) met on a weekly basis and operated using agile project management, conducting sprint planning, backlog refinement, and prioritization based on stakeholder needs and technical feasibility.

Additional smaller-scale, discipline-specific meetings (either in person or online) took place between therapists, sports scientists, and the development team. These focused on the collaborative creation and refinement of exercise routines, ensuring therapeutic relevance, patient safety, and age-appropriate design. Communication was supported through email and a shared digital workspace, which enabled asynchronous contributions and documentation.

\subsubsection{Embedded Testing and Feedback Loops}

At key development milestones, the robot system was evaluated in the actual deployment settings with relevant stakeholders, including nursing staff, therapists, and patients. These in-situ evaluations played a critical role in assessing acceptance, usability, and robustness under real-world conditions as well as training of therapists in using the system. 

The prototype robot system was adapted to additional use cases, including physiotherapy for pediatric surgery/urology in a hospital setting (InPT) and physical rehabilitation in an outpatient physiotherapy practice (OutPT). Each adaptation required re-evaluation of user needs, spatial constraints, and therapeutic goals, illustrating the non-linear, practice-driven nature of the co-development process.

\begin{table}[h!]
\centering
\renewcommand{\arraystretch}{1.2}
{\footnotesize
\begin{tabularx}{\textwidth}{>{\raggedright\arraybackslash}p{3.2cm} >{\raggedright\arraybackslash}p{3.5cm} >{\raggedright\arraybackslash}X >{\raggedright\arraybackslash}X}
\textbf{Collaborative Element (Frequency)} & \textbf{Actors} & \textbf{Activity} & \textbf{Outcome} \\
\hline

Full-team meetings (monthly) & Developers, physio and sports therapists, study nurse, psychologist, clinicians & 
Discuss organizational matters, 
demonstrate system state,
collect feedback and technical issues & 
Feedback on adapting system and implementing new system features, 
technical issues \\

\hline

Sprint meetings (weekly) & Developers (computer scientists) & 
Organize all feedback and technical issues into work items, 
prioritize work items, 
fill backlog for upcoming sprint & 
System increments \\

\hline

Quick exchanges (as needed) & Developers, physio and sports therapists & 
Show and discuss newly implemented exercises & 
Feedback for refinement of exercises \\

\hline

Workshop care staff children's center (once) & Developers, care staff & 
Demonstrate system state, collect feedback & 
Feedback on adapting system and implementing new system features \\

\hline

Visits to Deployment settings (quarter-yearly) & Developers, therapists, patients & 
Simulate use case, 
collect feedback, 
collect in-situ technical issues,
train therapists & 
Feedback,
technical issues \\

\end{tabularx}
}
\caption{Co-development process of Mobirobot}
\end{table}

\section{Resulting design principles}

\subsection{Robot Software} \label{sec:robot_software}



\subsubsection{Motivation}
\label{sec:motivation}

Children and adolescents, particularly those with depressive disorders in the psychiatric InST setting, often require additional motivation to actively participate in the sports program with Mobirobot. A workshop with children in which the robot was presented helped to spark interest in participating in the pilot study.

To support engagement, Mobirobot provides motivational phrases throughout the exercises. The phrases were chosen carefully to not evaluate the children's performance, which the robot could not perceive and to match both, exercises in sports and physiotherapy. Among them were utterances like [English translations] ``Great that we can do some exercises together!'', ``Keep it up!'', ``You're on your way to becoming a legend!'', etc. For participants with prior abdominal surgery, NAO also used phrases like ``Be careful!''.

In addition to these verbal prompts, we developed interactive buttons that children can use to play simple games with the robot. These buttons were used both during physiotherapy sessions at OutPT and as warm-up activities in the InST setting. The buttons emit light and sound, and connect to the laptop via Bluetooth. They can be flexibly placed within the exercise space, allowing Mobirobot to, for example, announce specific colors that the children are instructed to press. 
These small peripheral devices serve as engaging interaction tools that can help motivate children to participate more actively. They also hold potential to increase the long-term appeal of the program by introducing playful and varied elements.

\subsubsection{Interactivity}
We found an increased interactivity of the system to be important in all settings. This is in line with earlier more general findings emphasizing the importance of the interactivity of social robots for the engagement in rehabilitative therapies \citep{winkle2018social}. We addressed this need through motivational feedback and responsivity to the patient's own behavior, the possibility to chat with the robot, and increased expressivity (blinking).

For the InST setting, a pose recognition system based on OpenPose \cite{cao2019openposerealtimemultiperson2d} was implemented to enable the robot to detect the participants movements. The pose recognition used an external camera (cf. Section \ref{sec:monitor}). It identifies which part of the exercise the participant is performing based on detected joints from OpenPose. Based on this information, NAO can tailor its motivational prompts, for example, by encouraging engagement with phrases such as ``You can do one repetition'' when the participant is not actively participating in the training.

We began by training classifier models to detect specific poses within selected exercises. These models utilize joint position data obtained from OpenPose. For each exercise, we defined a set of distinct poses to facilitate detection of when the exercise was being performed. For instance, in the case of a squat, we categorized the movement into a \textit{standing} position and a \textit{squatting} position. When the classifier detects a transition from the standing to the squatting position, it is inferred that the participant has completed a repetition. The training data for these classification models were composed of recordings of adult individuals performing the predefined poses.

Statements from NAO based on time were added as well, e.g., ``There are 10 seconds left''. These personalized statements make the whole experience of training with NAO more interactive.
Another part that was added to make NAO more interactive is NaoChat.
NaoChat was developed as a dedicated state that NAO can enter before and after training, enabling the robot to respond dynamically during interactions with patients. The microphone is utilized to record the speech of the patient, which is subsequently converted into text using Whisper \cite{radford2022robustspeechrecognitionlargescale}. The text is entered into a compact LLM running on the laptop. For the LLM, Phi3 with 3.8B parameters was used. To support users in requesting further information about NAO, a basic overview of the robot and the project in which it is deployed has been provided. This includes information such as its name, physical dimensions and use-case in this project.

\begin{figure}
    \centering
    \includegraphics[width=.7\columnwidth]{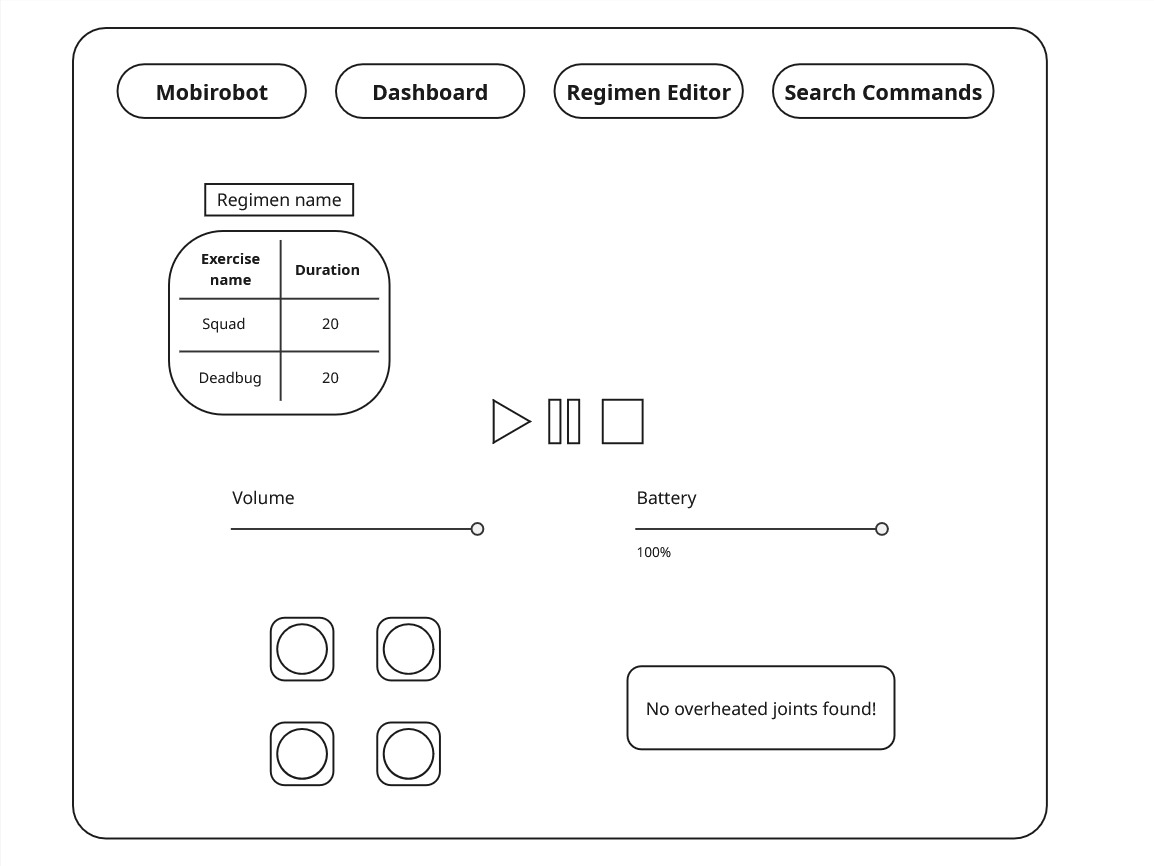}
    \caption{Schematic depiction of the GUI with regimen editor, controls and displays for system feedback.}
    \label{fig:new-gui}
\end{figure}

\subsubsection{Discreet monitoring and intervention}
\label{sec:monitor}
In the InST setting, children and adolescents with social anxiety responded positively to engaging with the robot alone, especially in contrast to the typical group-based therapy format. For safety reasons, however, therapists would like to be able to monitor the interaction and intervene if necessary. In our setting, children were not comfortable with an external camera visibly recording them. As a compromise, therapists monitored the sessions discreetly through a window in the room's door, maintaining both the children’s sense of privacy and the necessary level of supervision.
For physiotherapists, having control functions such as pausing, stopping, and resuming the exercises is essential. Given that Mobirobot is not yet capable of fully adaptive behavior, the ability to intervene manually remains crucial for addressing the individual needs and varying capabilities of patients.

\subsubsection{No-code feedback}
In addition to the above controls, providing no-code system feedback, such as visual or graphical indicators for battery status and potential overheating, enables therapists to quickly understand and respond to common technical issues without requiring specialized technical knowledge. The Bluetooth connection of the four buttons was also displayed.




\subsection{Exercise Program}
\label{sec:exprog}

\subsubsection{Adaptability}
\label{sec:adaptability}
One need we identified pertained to the adaptability of the training program in all settings. Therapists as well as patients want to adapt the system individually, rather than relying solely on a predefined program.
Children sometimes favored some exercises over others. To motivate them, children should have the option to customize their strength exercises according to their preferences and the intensity of their training, allowing them to tailor their workout before each session. In the physiotherapy settings, therapists should be able to adapt the exercises to the individual patient.

\begin{figure}
    \centering
    \includegraphics[width=1\linewidth]{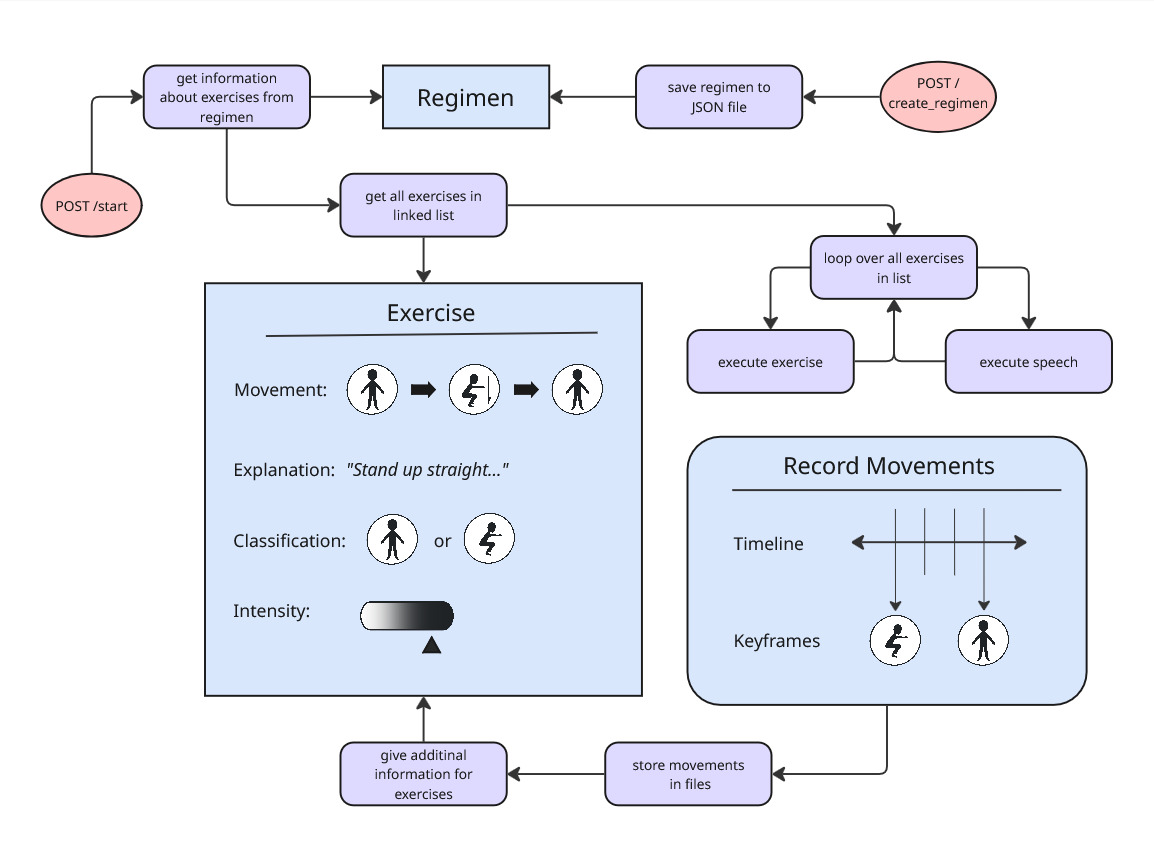}
    \caption{Overview of the flow of information in the backend}
    \label{fig:backend-overview}
\end{figure}

To be able to do so, so-called regimens were introduced. A regimen functions as a structured template for assembling new training programs. Regimens are groups of exercises, together with their duration, as well as the duration of the long and short breaks. The GUI enables the user to create regimens: Users can name each regimen and populate it with an arbitrary number of exercises arranged in the desired order. Users can customize the duration of the exercises and tailor both short and long break intervals according to their preferences. After a regimen is built, it is stored and can be used as often as the therapists want. An overview over how exactly a regimen is structured and stored can be found in Fig. \ref{fig:backend-overview}. 
Regimens can be started, paused, and stopped via the GUI. Once a regiment is started, the GUI displays which exercises will follow in the correct order. 
This part of the program can be accessed via a tablet or other internet capable device.

In addition to that, the GUI also allows the user to control certain parts of the robot (e.g., the volume, as well as to see information about the robot like the available battery and the temperature state of the joints) from outside the room in which the patient is exercising.  


The new and improved GUI can be found in Fig. \ref{fig:new-gui}.

\subsubsection{Hardware limitations: Balance, joint mobility and speed}

Several exercises such as abdominal breathing and various joint movements such as shoulder flexion/extension and elbow exercises were initially programmed, but were found to be unfeasible. Exercises like shoulder flexion/extension, which combined movements with breathing patterns, did not work because of NAO's technical limitations. 

Coordination exercises posed the greatest challenges. Many exercises were not operable because NAO could not shift its weight well enough, which is crucial in coordination exercises. Moreover, such exercises require fast execution, which was difficult to achieve because of NAO's technical characteristics. In the arm-twisting exercise, the robot could not raise its arms as high as needed. In the head-and-shoulders exercise, NAO was unable to perform the exercise as quickly as needed.

Included and excluded exercises for the individual settings can be found in the Supplementary Materials File 2.

Some exercises could not be demonstrated accurately by NAO due to limitations in balance, restricted joint mobility, and resulting unstable or slow movements. Nevertheless, these constraints can be addressed through appropriately designed compensatory verbal cues, which acknowledged NAO’s physical limitations and instruct children to instead perform the exercises correctly themselves, for example, to perform a movement faster or more precisely than NAO could demonstrate.

\subsection{Implementation in the real-world medical settings}
\subsubsection{Setting characteristics: Setup and Floor Structure}
Across the different deployment settings, several adaptations were made to accommodate both environmental and therapeutic needs. In the pediatric surgery/urology ward (InPT), while in use, Mobirobot was transported between rooms by therapists and stored in accessible locations to allow for quick deployment. Because children recovering from abdominal surgeries remained in bed, Mobirobot was placed on a mobile trolley to maintain visibility and enable exercise demonstration at eye level. In OutPT sessions for children recovering from lower-limb fractures, Mobirobot was positioned on the floor of the practice’s gym, similarly ensuring a clear line of sight. In contrast, in the InST context, Mobirobot operated in a designated sports hall during one-on-one sessions. Here, movement routines were specifically programmed to suit the local floor structure.
When doing the internal test run, it was found that NAO often lost balance during the exercise `balancing'. NAO needs a very even floor structure in order to be able to move properly on it. These conditions could not always be met, which led to problems during the training sessions. By replacing this exercise with exercises in which NAO is more stable the sport program could still be carried out in full.
Another issue concerned standing up after the floor exercises. On the floor in the InST setting, Mobirobot constantly turned a bit to the side when standing up and, after several exercises, ended up standing with its back to the subject. This showed that a position correction was needed after the floor exercises. Through several tests, it was observed to which side and by how much Mobirobot turns when standing up. Based on these observations, a quick walking correction was implemented to allow Mobirobot to return to its starting position after standing. The exercises for robot-assisted physiotherapy settings were developed following the same 3-step process as the development of the exercise program prototype described in Section~\ref{sec:exptototype} and the Supplementary Materials File 1.

\subsubsection{Simple setup}
The prototype setup included a laptop, a tablet, the robot, a Wi-Fi router, an external camera on a tripod, and a smart watch for heart rate monitoring.
To reduce the burden on staff, the deployment was designed to minimize equipment and procedural complexity. For instance, the heart rate monitoring functionality, initially included, was discontinued due to recurring technical issues and limited reliability. The use of a dedicated Wi-Fi router was rendered unnecessary by employing a laptop that could establish direct Wi-Fi connections to the peripherals. These decisions aligned with the overarching goal of keeping the setup simple and robust, particularly in environments with varying infrastructural conditions.

Due to discomfort expressed by patients, the external camera for pose recognition is only employed when approved by the patients.

\subsubsection{Training of staff}
In addition to simplifying the technical setup, considerable emphasis was placed on training the clinical staff to ensure confident and autonomous use of the system. This training occurred both informally during the development phase,through close collaboration and observation, and in structured sessions designed to build familiarity with the robot and the GUI. Therapists were given the opportunity to practice independently with Mobirobot, both in the university lab and in real-world clinical environments, allowing them to simulate and rehearse use-case scenarios under realistic conditions. To further support autonomous usage, printed and digital (on the laptop) documentation was provided, including a detailed step-by-step guide, a list of frequently asked questions, and troubleshooting instructions for resolving common technical issues. This multi-tiered training approach was essential not only for ensuring operational stability across diverse settings, but also for strengthening therapists’ confidence in adapting the system to their therapeutic routines.

\subsubsection{Remote Servicing}

Maintaining Mobirobot in active clinical and outpatient settings requires the ability to provide remote servicing and support. This proved to be both a critical capability and a recurring challenge, particularly given the inconsistent availability of stable internet connections in some hospital environments. Remote desktop sessions are scheduled whenever local issues arise that could not be resolved by the clinical team. These sessions are primarily used to address system-level problems outside the scope of our application, such as switching WLAN configurations or managing operating system updates. In rare cases, remote access is also employed to observe and reconstruct errors that have occurred on-site but could not be reproduced during local testing. While tools for remote access and control are used for establishing remote connections, these interventions are heavily dependent on a functioning and secure internet connection, without which diagnosis and repair are delayed. The importance of reliable connectivity for remote servicing cannot be overstated: it forms a prerequisite for effective troubleshooting and continuous system operation, yet remains one of the least predictable aspects in the field.

\section{Discussion}

This paper presented the development and early-stage deployment of Mobirobot, a socially assistive robot designed for paediatric therapeutic contexts. The project was guided by a multidisciplinary and interprofessional team using agile and human-centred design principles. Our work builds upon a prior single-setting prototype, advancing it into a flexible platform tailored for diverse clinical and therapeutic environments. Emphasis was placed on early real-world integration within paediatric medical settings, allowing iterative technical and usability refinements through close collaboration with healthcare professionals.

The development process foregrounded inclusive co-design, ensuring responsiveness to the needs and constraints of actual users. Based on these insights, the system was iteratively improved to better support therapeutic goals in real-life practice.

Our findings reinforce prior claims that human-centred and co-creative robotic design is crucial for clinical uptake and effectiveness \citep{muller2024cocreative}. The iterative development of Mobirobot echoes the participatory robotics paradigms advocated in earlier work on therapeutic social robots \citep{winkle2018social, olatunji2024immersive}. However, this project adds a distinctive contribution by incorporating agile methods and deploying early in real-world clinical environments, rather than relying solely on lab-based development or late-stage field trials.
Notably, while prior systems such as KASPAR~\citep{wood2021developing,dautenhahn2009kaspar,robins2018kaspar} and Probo~\citep{saldien2008design,vanderborght2012using} have focused predominantly on autism interventions or home-like settings, Mobirobot addresses a broader paediatric patient base, including those undergoing surgery or psychiatric treatment.

We are currently conducting a feasibility study to assess Mobirobot's acceptability and practical implementation in InPT, OutPT and InST. The study evaluates two primary aims: (i) to explore the usability, perceived benefit, and interaction quality from the perspectives of both patients and staff; and (ii) to understand the implementation needs, acceptability as well as engagement and therapy potential of the robot in acute inpatient and outpatient settings. Data collection includes standardised and exploratory questionnaires on satisfaction and acceptance, alongside behavioural observations and post-session interviews. Challenges in participant recruitment, particularly in psychiatric settings (InST), are ongoing and have informed the adaptive study design.

\subsection{Strengths and Limitations}
A major strength of this project is its commitment to stakeholder-centred design and implementation. Throughout the development process, healthcare professionals from multiple disciplines---including child psychiatry, surgery, nursing, and physical therapy---were continuously involved. Their feedback informed not only the interaction scenarios but also hardware configurations, safety concerns, and ethical considerations. The resulting co-development strategy enabled context-sensitive deployment and increased the likelihood of successful implementation.

Particularly, the adaptability of Mobirobot across different clinical environments was only achievable through this collaborative framework. For example, robot behaviours were adapted to suit both open ward and private therapy-room settings, addressing constraints such as movement restrictions, patient receptivity, and clinical workflow integration. This supports the growing consensus that sustainable digital health interventions must evolve in partnership with users across the full lifecycle of the system.

Across sites, children were generally very happy to interact with the robot, and an illustrative case underscores this effect: a shy boy, reportedly perhaps on the autism spectrum (undiagnosed), completed the full set of exercises when Mobirobot was present; as his mother noted, “had the robot not been there, he probably wouldn’t have done the physiotherapy at all.” 

Several limitations emerged during development and early evaluation. First, achieving long-term engagement with the robot remains an open challenge, particularly in psychiatric settings where patients' willingness to interact may fluctuate. Second, while the current version includes basic verbal interaction and movement, further interactivity modules, such as engagement detection~\citep{loos2024decoding}, affect recognition~\citep{rouast2019deep}, or pain monitoring for children, are not yet implemented. Third, technical limitations persist. The lack of an external camera simplifies logistics and enhances patient comfort, but compromises the system’s ability to detect fine-grained engagement cues or maintain robust pose recognition. Adding to this challenge, especially in contexts like surgery or urology, bed-ridden patients are partially occluded. While the NAO platform provides basic onboard sensing, certain activities require additional sensory input because the platform and its sensors are moving during exercises.
Finally, patient recruitment for the ongoing study has been slower than anticipated. Reasons include the burden of consent procedures, especially in psychiatry, and competing clinical priorities. Adaptive adjustments to the recruitment strategy and inclusion criteria are being explored to mitigate these issues.


\section{Conclusion}
This paper outlines the multiprofessional, iterative co-development of Mobirobot, a socially assistive robot designed to support mobilisation in paediatric patients recovering from trauma or fractures and in children and adolescents with depressive disorders. By demonstrating and performing exercises alongside patients, Mobirobot fosters engagement in therapeutic movement. The system includes adaptable exercise routines with interactive explanations, motivational prompts, and a graphical user interface for monitoring progress and providing no-code feedback. Its accessible setup and focus on user-centred design make it particularly suited to clinical integration.

Through an agile, stakeholder-driven development process, we identified overarching feedback themes that hold relevance beyond this project and may guide future socially assistive robotics in healthcare. These insights highlight the importance of embedding real-world constraints, enabling co-adaptation, and fostering usability through transparency and simplicity in robot behaviour.

While this paper has critically addressed technical limitations, particularly regarding sensing and hardware, we also wish to acknowledge the strength and reliability of the NAO platform. The robust performance of the hardware across repeated use in the lab and the clinical environments is a testament to the engineering by Aldebaran and the NAO development team.

Taken together, our findings demonstrate that a stakeholder‑centred, context‑aware development process can yield a practically deployable robotic system with the potential to enrich paediatric rehabilitation and mental‑health interventions, while offering clear design principles for future therapeutic robot development.

\subsection{Ethical considerations}
The development process involving human participants was reviewed and approved by the ethics committee Ethik-Kommission Westfalen-Lippe (2023-282-fS). The patients/participants provided their written informed consent to participate in this study. 

\section*{Conflict of Interest Statement}
The authors declare that the research was conducted in the absence of any commercial or financial relationships that could be construed as a potential conflict of interest.

\section*{Author Contributions}
Conceptualization: LD, AG, MN, ALR, JB, JM, MH, AB, JL, BKZ, NF, AV, MM, IKP, MS, WB, ALV, Project administration: ALV, NF, JM, LD, Validation: LD, AG, MN, ALR, JB, JM, MH, AB, JL, BKZ, NF, AV, MM, IKP, MS, WB, ALV, Investigation: LD, AG, MN, ALR, JB, JM, MH, JA, AB, JL, BKZ, MC, NF, AV, MM, IKP, MS, WB, ALV, Resources: ALV, WB, MS, BKZ, JS, Funding acquisition: ALV, WB, Software: LD, AG, MN, ALR, Supervision: ALV, WB, MS, BKZ, JS, Writing---original draft: ALV, AB, JM, JA, MM, Writing---review \& editing: all. For definitions see http://credit.niso.org/ (CRediT taxonomy). All authors approved the submitted version.

\section*{Funding}
We gratefully acknowledge the financial support from the Medical School OWL for the project Mobirobot.

\section*{Acknowledgments}
We acknowledge the contributions of specific colleagues, institutions, or agencies that aided the efforts of the authors.

\section*{Data Availability Statement}
The meeting protocols and Mobirobot code base are available from the corresponding author upon reasonable request..

\section*{Supplemental Material}

\section{File 1: Exercise review process}
\label{suppmat1}

\begin{table}[h!]
\centering
\renewcommand{\arraystretch}{1.2}
\definecolor{stepgray}{gray}{0.9}
\begin{tabularx}{\textwidth}{@{}l X c@{}}
\toprule
\multicolumn{3}{@{}l}{\textbf{Step 1: Exercises were selected, classified and programmed for Mobirobot}} \\
\midrule
\textbf{Domain} & \textbf{Exercise Types} &  \\
\midrule
Inpatient sports therapy (InST) & Strength, Coordination, Stretching &  \\
Inpatient physiotherapy (InPT) & Arm, Leg, Foot, Breathing & \\
Outpatient physiotherapy (OutPT) & Mobilization, Strengthening, Stretching & \\
\midrule
\multicolumn{3}{@{}l}{\textbf{Step 2: First test of feasibility through an internal test run}} \\
\midrule
\textbf{Domain} & \textbf{Exercise Type} & \textbf{Fulfilled / Total} \\
\midrule
\textbf{InST} & Strengthening & 14 / 18 \\
                                         & Coordination & 0 / 4 \\
                                         & Stretching & 4 / 4 \\
\midrule
\textbf{InPT} & Arm & 6 / 8 \\
                                       & Leg & 3 / 5 \\
                                       & Foot & 3 / 4 \\
                                       & Breathing & 1 / 2 \\
\midrule
\textbf{OutPT} & Mobilization & 8 / 8 \\
                                 & Strengthening & 12 / 15 \\
                                 & Stretching & 5 / 7 \\
\midrule
\multicolumn{3}{@{}l}{\textbf{Step 3: Second test of feasibility through an internal test run}} \\
\midrule
\textbf{Domain} & \textbf{Final number of exercises} & \\
\midrule
\textbf{InST} & 18 & \\
\textbf{InPT} & 13 & \\
\textbf{OutPT} & 25 & \\
\bottomrule
\end{tabularx}
\caption{Stepwise Feasibility Testing and Classification of Exercises for Mobirobot}
\end{table}


\section{File 2: Categorized Exercise List by Setting}
\label{suppmat2}

\subsection{Inpatient sports therapy (InST)}

\subsubsection*{Included Exercises}

\textbf{Strength:} 
\begin{itemize}
\item Boxing
\item Push-up Variations
\item Sit-Up
\item Plank
\item Hand Push
\item Lunge
\item Dead Bug
\item Jackknife
\item Leg Raise
\item Russian Twist
\item Superman
\item Hacker
\item Hip Raise
\item Squat
\end{itemize}

\textbf{Stretching:}
\begin{itemize}
\item Side Stretch
\item Neck Stretch
\item Trunk Stretch
\item Calf Stretch
\end{itemize}

\subsubsection*{Excluded Exercises}

\begin{itemize}
\item Rotate arms in opposite directions
\item Left arm makes large circles, right arm makes small circles, and vice versa
\item Stand on one leg, bring arms overhead to touch, then switch the standing leg
\item Half jumping jack, arms straight up while legs alternate left/right
\item Head, Shoulders, Hips, Knees, Ankles, possibly without crossing
\item Stand on one leg, look down, up, left, right, then switch the standing leg
\item Finger Exercise
\item Hare/Hunter
\end{itemize}

\subsection{Inpatient physiotherapy (InPT)}

\subsubsection*{Included Exercises (in a lying position)}

\textbf{Arm:}

\begin{itemize}
\item Apple picking
\item Arms stretching Variations
\item Boxing
\item Elbow Variations
\item Hand fist
\item Wool wrap
\end{itemize}

\textbf{Leg:}

\begin{itemize}
\item Legs outside Variations
\item Bend leg right/left
\item Hip Raise Variations
\end{itemize}

\textbf{Foot:}

\begin{itemize}
\item Pull back your feet Variations
\item Moving feet Variations
\item Feet circling Variations
\end{itemize}

\textbf{Breathing:}

\begin{itemize}
\item Abdominal breathing
\end{itemize}

\subsubsection*{Excluded Exercises}

\begin{itemize}
\item Nasal breathing
\item Holding the most painful wound during movement (collision risk)
\item Isometric tension exercises of the whole legs
\item Toe curling (NAO does not have toes)
\item Finger games (NAO has only three fingers)
\item 90-degree shoulder flexion followed by crossing the arms (collision risk)
\end{itemize}

\subsection{Outpatient physiotherapy (OutPT)}

\subsubsection*{Included Exercises}

\textbf{Mobilization:}

\begin{itemize}
\item Legs outside
\item Leg outside right/left
\item Bend leg right/left
\item Pull back your feet
\item Pull back your foot right/left
\item Moving feet
\item Feet circling
\item Foot circling right/left
\end{itemize}

\textbf{Strengthening:}

\begin{itemize}
\item Dead Bug
\item Hip Raise Variations
\item Jackknife
\item Leg Raise
\item Lunge
\item Plank
\item Push Up Variations
\item Sit Up Variations
\item Squat
\item Side hole squats
\item Superman
\item Crawling football
\end{itemize}

\textbf{Four-footed stand:}

\begin{itemize}
\item Four-footed kick right/left
\item Four-footed outside right/left
\item Four-footed bend right/left
\item Four-footed stretching right/left
\end{itemize}

\textbf{Stretching:}

\begin{itemize}
\item Stretching the calf muscles in a lunge position
\item Stretching the adductors; one leg to the side, the other bent at the knee joint
\item Stretching the hip flexors in a lunge position
\item Stretching the quadriceps
\item Ischios
\end{itemize}

\subsubsection*{Excluded Exercises}

\begin{itemize}
\item Squats with lateral steps (balance)
\item Crawl football (balance in asymmetry)
\item Pressing buzzers with feet while standing (balance on one foot)
\item Hip flexor stretch in lunge position (balance and joint limits)
\item Singing the nursery rhyme "Aramsamsam" while stretching
\end{itemize}

\bibliographystyle{Frontiers-Harvard}
\bibliography{refs}

\end{document}